\documentclass[lettersize,journal]{IEEEtran}
\usepackage{amsmath,amsfonts}
\usepackage{algorithmic}
\usepackage{algorithm}
\usepackage{array}
\usepackage{comment}
\usepackage{color}

\usepackage[caption=false,font=normalsize,labelfont=sf,textfont=sf]{subfig}
\usepackage{textcomp}
\usepackage{stfloats}
\usepackage{url}
\usepackage{verbatim}
\usepackage{graphicx}
\usepackage{cite}
\begin{document}

\title{Multimodal Remote Sensing Scene Classification Using VLMs and Dual-Cross Attention Networks}

\author{Jinjin Cai$^1$, Kexin Meng$^1$, Baijian Yang$^1$, and Gang Shao$^2$ 

\thanks{$^1$Department of Computer and Information Technology, Purdue University, West Lafayette, IN 47907, USA. {\tt\small{[cai379, meng147, byang]@purdue.edu}.}}
\thanks{$^{2}$School of Information Studies, Purdue University, West Lafayette, IN 47907, USA. {\tt\small{gshao@purdue.edu}.}}

}

\markboth{Journal of \LaTeX\ Class Files,~Vol.~14, No.~8, August~2021}%
{Shell \MakeLowercase{\textit{et al.}}: A Sample Article Using IEEEtran.cls for IEEE Journals}

\IEEEpubid{0000--0000/00\$00.00~\copyright~2021 IEEE}

\maketitle

\begin{abstract}
Remote sensing scene classification (RSSC) is a critical task with diverse applications in land use and resource management. While unimodal image-based approaches show promise, they often struggle with limitations such as high intra-class variance and inter-class similarity. Incorporating textual information can enhance classification by providing additional context and semantic understanding, but manual text annotation is labor-intensive and costly. In this work, we propose a novel RSSC framework that integrates text descriptions generated by large vision-language models (VLMs) as an auxiliary modality without incurring expensive manual annotation costs. To fully leverage the latent complementarities between visual and textual data, we propose a dual cross-attention-based network to fuse these modalities into a unified representation. Extensive experiments with both quantitative and qualitative evaluation across five RSSC datasets demonstrate that our framework consistently outperforms baseline models. We also verify the effectiveness of VLM-generated text descriptions compared to human-annotated descriptions. \textcolor{black}{Additionally, we design a zero-shot classification scenario to show that the learned multimodal representation can be effectively utilized for unseen class classification. This research opens new opportunities for leveraging textual information in RSSC tasks and provides a promising multimodal fusion structure, offering insights and inspiration for future studies.} Code is available at: \url{https://github.com/CJR7/MultiAtt-RSSC}
\end{abstract}


\section{Introduction}\label{INTRODUCTION}
\IEEEPARstart{R}{emote} sensing scene classification (RSSC) \cite{thapa2023deep} is a critical task in analyzing satellite or aerial imagery, where spectral representations are classified into predefined land cover classes or scene types. This task is vital for converting raw remote sensing data into actionable information, enabling various applications such as environmental monitoring \cite{oghaz2019scene}, land use analysis \cite{fang2022spatial}, urban planning \cite{su2021urban}, and image retrieval \cite{gu2019survey}, among others. 

Historically, traditional methods for land cover classification have relied heavily on image processing techniques and machine learning models. These conventional approaches often utilize hand-crafted features, such as color histograms (CH) \cite{dos2010evaluating}, histograms of oriented gradients (HOG), and scale-invariant feature transform (SIFT) \cite{luo2013indexing,sheng2012high}. These features are carefully designed by domain experts, targeting specific aspects of the image. However, their ability to effectively represent global features is limited, reducing their performance and applicability in complex classification tasks.
\begin{figure*}[t]
\begin{center}
\includegraphics[width=1.0\textwidth]{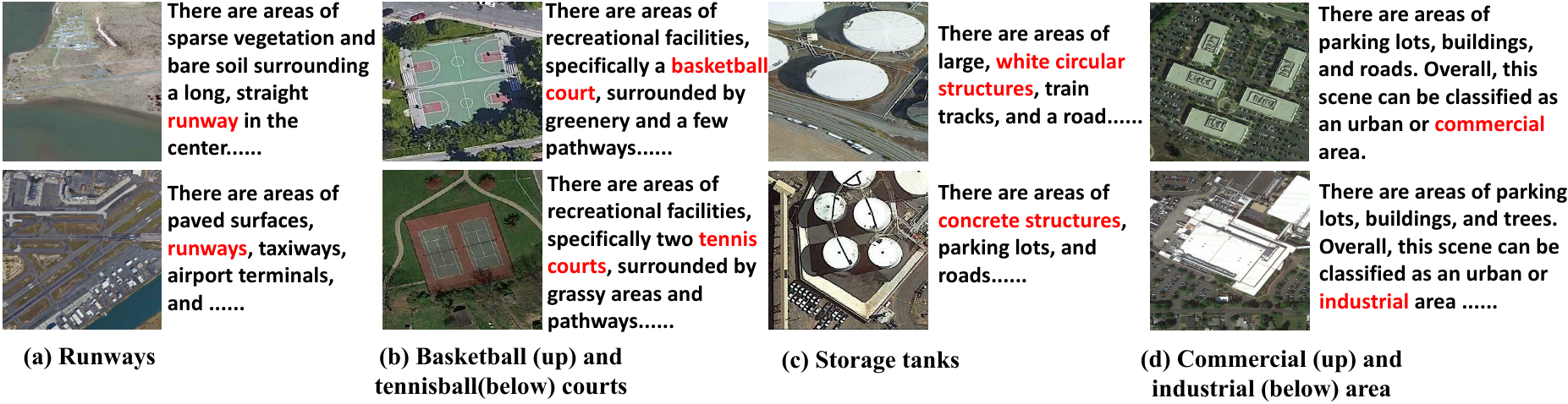}
\caption{Examples of challenges in image-only RSSC methods and zero-shot descriptions generated by the LLaVa VLM: (a) High intra-class variance; (b) High inter-class similarity; (c) Large variations in the scale of objects or scenes; (d) Coexistence of multiple ground objects within a single image.}
\label{fig:problem}
\end{center}
\end{figure*}
The emergence of deep learning has significantly transformed the scene classification landscape, enabling models to autonomously learn and extract meaningful features from data using neural networks \cite{cheng2020remote}. Leading deep learning models in this domain include Convolutional Neural Networks (CNNs) \cite{liu2018scene}, Generative Adversarial Networks (GANs) \cite{guo2020gan}, and Visual Transformers (ViTs) \cite{lv2022scvit,bi2022vision}. 

Despite these advancements, image-based unimodal scene classification still faces several challenges. As Cheng et al. \cite{cheng2020remote} have noted, these challenges include high intra-class variance, where images within the same class may appear vastly different, and high inter-class similarity, where images from different classes can look strikingly similar. Additionally, there are large variations in the scale of objects or scenes, as well as the coexistence of multiple ground objects within a single image. To effectively address these challenges, fine-grained image analysis \cite{shendryk2018deep} and sophisticated feature extraction \cite{liu2023multi} methods are typically required. However, the availability of such high-resolution and detailed images is often limited \cite{he2018high}.

A promising approach to overcoming these limitations is the incorporation of text as an additional modality, providing a rich source of complementary information. Text captions or descriptions have been successfully leveraged in various tasks such as scene retrieval \cite{yuan2021lightweight}, image captioning \cite{wang2022multi}, and few-shot/zero-shot classification \cite{cheng2023teaw}. Textual descriptions can capture detailed information in remote sensing images that may be challenging to recognize from the visual data alone. By complementing visual features, these textual descriptions can highlight abstract socio-semantic attributes of the scene, enriching the overall understanding. Moreover, integrating text with images can also enhances the interpretability of the classification process, providing us a deeper understanding of the rationale behind classification decisions.

\textcolor{black}{Despite growing interest in integrating textual data into remote sensing tasks, significant challenges persist, primarily due to the difficulty of acquiring large volumes of high-quality and rich descriptive text.} Currently, remote sensing (RS) text description (image captioning) datasets are primarily created through manual or automated annotation processes, both of which have limitations: \textcolor{black}{On the one hand, the manual generation of descriptions is labor-intensive. Annotating RS images with detailed text descriptions is extremely time-consuming and requires significant expertise in various land cover types, environmental factors, and even local geographic knowledge. Meanwhile, the manually generated descriptions may be incomplete, because annotators are more likely to focus on the main elements of the image, potentially overlooking subtle details or important background information. Additionally, variations in descriptive habits among different annotators can lead to inconsistencies in annotation formats. On the other hand,} automated methods for generating descriptions, while more efficient, can also miss important information. For example, in a recent study \cite{liu2024remoteclip}, the authors employed ``mask to box (M2B)" and ``box to caption (B2C)" approaches to convert masks or detection box labels from remote sensing segmentation datasets into image-text pairs. However, this approach fails to describe the positional relationships of objects within the image and the background objects that are not labeled in the segmentation dataset.

The emergence of advanced vision-language models (VLMs), such as LLaVa and ChatGPT-4-V, appears promising in addressing these challenges. These models can generate detailed descriptions as a zero-shot source for providing textual information. As illustrated in Fig.~\ref{fig:problem}, textual descriptions generated by VLMs from RS images can highlight subtle visual objects or contextual information that may be overlooked by image classification models. For example, they can specify the density of buildings or the presence of specific landmarks, thereby providing complementary details.
However, the potential of VLMs is contingent on the development of efficient multimodal fusion approaches. While text and visual data are inherently complementary, simply combining them using basic strategies, such as concatenation or late fusion, can introduce significant challenges. These straightforward fusion methods often fail to capture the complex interdependencies between modalities, potentially leading to the addition of noise rather than the enhancement of the model performance. This noise can dilute the informative value of either modality, especially in the context of remote sensing, where the information is often subtle and distributed across both visual and textual domains. This challenge is particularly pronounced when dealing with text generated by VLMs, as these models are prone to AI hallucinations, where the generated text may not accurately reflect the visual content or may introduce inaccuracies. Such inconsistencies can lead to unreliable or misleading multimodal representations if not properly addressed.

Inspired by these challenges, in this work, we propose a multimodal framework that leverages text descriptions generated by VLMs as an auxiliary modality to enhance RSSC. We introduce a dual cross-attention module designed to capture the intricate dependencies between visual and textual data, effectively leveraging the complementary information provided by each modality. This approach ensures a robust fusion of modalities, thereby improving classification accuracy and interpretability. Our key contributions are summarized as follows:
\begin{itemize}
    \item We advocate for incorporating VLM-generated text descriptions as a multimodal auxiliary input to enhance remote sensing scene classification (RSSC) without the need for extensive manual annotation. 
    \item We introduce a dual cross-attention module specifically designed to capture the intricate dependencies between visual and textual data, enabling a more effective fusion of modalities and robust representation learning \textcolor{black}{with flexibility in VLM and modality-wise encoder choices.}
    \item We conducted extensive experiments across five datasets and compared seven baseline models to demonstrate the effectiveness and robustness of our proposed framework in diverse scenarios and data conditions. We also provide visualizations to enhance the interpretability of the results, highlighting how our model captures and leverages multimodal information. 
    \item We further explored the potential of multimodal representation in zero-shot classification tasks, demonstrating its ability to better bridge the gap between seen and unseen classes.

\end{itemize}

\section{Related work}\label{sec:RELATED WORK}

\subsection{Remote Sensing Scene Classification (RSSC)}
Remote sensing scene classification involves categorizing satellite or aerial imagery into predefined land cover classes, transforming raw remote sensing data into actionable insights for applications such as urban planning, environmental monitoring, and disaster management. Recent advancements in deep learning have introduced new approaches to RSSC. For example, Convolutional Neural Networks (CNNs) have demonstrated significant potential due to their ability to extract deep features and manage irregular spatial dependencies in images. Several CNN architectures have been adapted and optimized for RSSC, including ResNet \cite{tong2020land}, DenseNet \cite{shen2021dual}, Inception Networks \cite{wang2018scene}, and EfficientNet \cite{alhichri2021classification}. For example, Alhichri \textit{et al.} introduced RS-DeepSuperLearner, an ensemble method that fuses outputs from five advanced CNN models to optimize classification performance \cite{alhichri2023rs}. In another notable work, Hou \textit{et al.} proposed an end-to-end network utilizing a pretrained ResNet34 to extract multi-level features, alongside a Contextual Spatial-Channel Attention Module (CSCAM) to generate attention features by exploiting information from different network levels \cite{hou2023contextual}.

More recently, Vision Transformers have emerged as a promising alternative in RSSC, particularly for their ability to capture long-range dependencies and global context in images. These models address some of the limitations inherent in CNN architectures, especially in handling complex spatial relationships across large image areas. For instance, Peng \textit{et al.} proposed a Local-Global Interactive Vision Transformer (LG-ViT) that effectively learns the interaction and constraints of both local and global features, thereby tackling challenges posed by dramatic variations in scale and object size typical in RSSC tasks \cite{peng2023local}.


While these advancements have significantly advanced RSSC performance, challenges persist, particularly in managing the high intra-class variance and inter-class similarity often found in remote sensing imagery. Therefore, integrating information from other modalities, such as textual descriptions, emerges as a promising direction for further improving RSSC performance.

\subsection{Multimodal Fusion and Text Modality in RSSC tasks}

Multimodal approaches have emerged as a promising direction in RSSC by integrating complementary information from various data types to enhance classification accuracy, especially in challenging cases. Recent applications of multimodal learning in RSSC have explored the fusion of visual data with other modalities, such as spectral information \cite{soroush2023nir}, Light Detection and Ranging (LiDAR) data \cite{liu2021mrssc,ma2023optical}, and temporal data \cite{sun2024alignment}. \textcolor{black}{While these multimodal approaches show promise, the acquisition and processing of such data require specialized equipment and expertise, making these methods less accessible for large-scale or resource-constrained applications.} 

\textcolor{black}{As an alternative, the use of textual information extracted from images represents a cost-effective and scalable modality, which provides semantic context that can enhance RSSC tasks without the need for specialized sensors. Textual information has been widely used in remote sensing to guide tasks such as semantic segmentation \cite{zhang2023text2seg} and image-text retrieval \cite{yuan2023parameter,tang2023interacting}. However, text annotation in remote sensing also presents challenges. On the one hand, manual annotation is not only labor-intensive but also requires specialized knowledge of land cover types, environmental factors, and geographic details. This process is time-consuming and prone to omissions, as annotators tend to focus on prominent elements in the image, often neglecting subtle details or background information. Furthermore, variations in descriptive habits among annotators can result in inconsistent annotation formats. On the other hand, automated text generation from remote sensing segmentation or detection datasets offers a more efficient alternative but is not without limitations \cite{liu2024remoteclip}. These approaches fail to capture positional relationships between objects and overlook unlabeled background elements, reducing the comprehensiveness of the descriptions. As a result, the use of text modality in RSSC remains underexplored compared to sensor-based fusion approaches.} 

\subsection{Foundation/Vision-Language Models}

Recent advancements in Vision-Language Models (VLMs) have opened new avenues for enhancing RSSC tasks by integrating textual information. These models, trained on vast datasets of image-text pairs, excel at understanding and generating natural language descriptions of visual content. While VLMs have seen limited use in the remote sensing domain \cite{qiu2024few}, their success in fields like medicine, computer vision, and social media highlights their potential. For instance, studies in medicine \cite{byra2023few,ji2024vision,qin2022medical,zhong2024vlm} have demonstrated VLMs' ability to describe medical images across modalities, improving interpretation and classification.

In remote sensing, differentiating between fine-grained land cover types and objects with varying scales presents a unique challenge. Several works \cite{saha2024improved,tzelepi2024exploiting} have explored VLMs for fine-grained image classification in other domains, which is particularly relevant for RSSC, where distinguishing between similar classes is crucial. Additionally, VLMs can provide abstract contextual information—such as historical land use or cultural significance—that may not be directly visible in the imagery.

Our work builds upon these advances but pushes the boundary further by incorporating VLM-generated text descriptions directly into the RSSC framework, providing a novel solution to enhance classification accuracy without the high costs associated with manual text annotation. Unlike previous approaches in other domains, we propose a dual cross-attention mechanism to fuse visual and textual data, ensuring robust integration of both modalities. This fusion allows our model to effectively leverage the complementary strengths of images and text, improving scene classification and overcoming the limitations of zero-shot VLM text generation, such as hallucinations.

\begin{figure*}[h]
\begin{center}
		\includegraphics[width=1.0\textwidth]{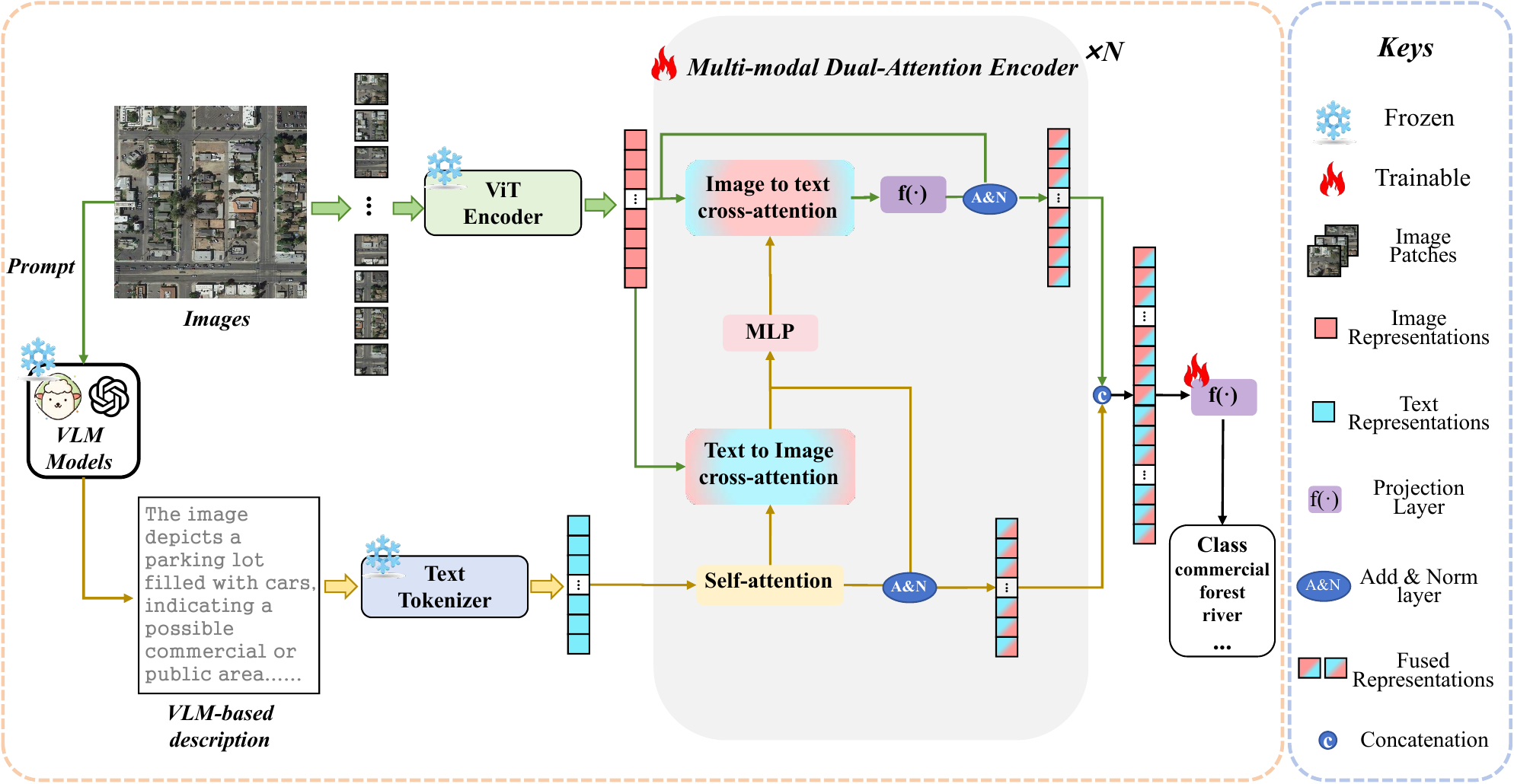}
	\caption{Illustration of the proposed framework. The RS images are paired with designed prompts and input into VLMs to generate text descriptions. These images are then processed by the Image Encoder to extract image embeddings, while the VLM-generated text is tokenized using the Text Encoder to produce text embeddings. Both sets of embeddings are fed into the Multimodal Dual-Attention Encoder where the complementary information are integrated and enhanced. The fused multimodal representations are then passed through a projection layer to produce the final classification output.
}
\label{fig:overview}
\end{center}
\end{figure*}

\section{Proposed method}\label{sec:PROPOSED METHOD}
In this section, we present our proposed VLM-based multimodal dual-attention framework for remote sensing scene classification. As illustrated in Fig.~\ref{fig:overview}, the network comprises three core modules: Visual Encoder, Text Tokenizer, and Multimodal Dual-Attention Encoder. \textcolor{black}{Our goal is to validate the effectiveness of the proposed multimodal structure, and the Visual Encoder and Text Tokenizer can be replaced with other pre-trained models. In this study, we use Vision Transformer as the Visual Encoder and CLIP's Text Tokenizer as examples.}

The process begins with images from the scene classification dataset, which are paired with designed prompts and input into VLMs to generate text descriptions. It is important to note that the VLM is utilized in its pretrained state solely for generating high-quality textual descriptions and is not fine-tuned during this process.

Our approach leverages ViTs for image encoding and the CLIP  text encoder for text encoding. Specifically, images are input into the ViT to extract image embeddings, while the text generated by the VLM is processed by the Text Tokenizer (CLIP text encoder) to produce corresponding text embeddings.

These image and text embeddings are then passed through N layers of the Multimodal Dual-Attention Encoder. This encoder employs attention mechanisms to effectively integrate and enhance the complementary information from both modalities, resulting in a robust and enriched representation of the scene. The enriched multimodal representation is subsequently fed into a prediction layer, which produces the final classification decision.

In the following sections, we will delve into the technical details of each module.

\subsection{VLM-based Description Generation}
For generating VLM-based descriptions, we utilized LLaVA (Large Language and Vision Assistant), an open-source, end-to-end large-scale multimodal model \cite{liu2024visual}. \textcolor{black}{The choice of the LLaVA in this study was motivated by its public availability. However, the framework is designed to be agnostic to the specific VLM used. Other VLMs, such as BLIP, Flamingo, or GPT-based multimodal extensions, could also be integrated without significant changes to the structure.} In detail, LLaVA integrates visual encoders with large language models (LLMs), achieving a generalized understanding of both vision and language. This integration has demonstrated excellent performance in tasks such as visual question-answering, reasoning, and content recognition. In our application, LLaVA is employed to generate comprehensive descriptions of remote sensing images based on carefully crafted prompts.

We observed that using LLaVA-referable templates within the prompts significantly increased the relevance and accuracy of the generated descriptions, leading to improved experimental outcomes. After extensive testing, we designed the following prompt for our application:

\texttt{This is an aerial image showcasing a particular land cover type. From a scene classification perspective, describe the content in the image as concisely as possible, focusing on the main features and overall classification.}

\texttt{Sentence Template: ``There are areas of {land cover type 1}, {land cover type 2}, {additional objects or features as needed}. Overall, this scene can be classified as [primary land cover classification] or {secondary land cover classification if needed}."}

To ensure that the generated descriptions are based solely on the visual content of the images and not influenced by external cues, we renamed all image files in our dataset using a combination of numbers. This approach prevents any unintended hints or biases that could arise from descriptive filenames. Examples of the descriptions generated by LLaVA can be found in Fig.~\ref{fig:problem}.

\subsection{ViT Image Encoder}

We utilize hidden layers of pretrained ViT as the image encoder in our model \cite{dosovitskiy2020image}, \textcolor{black}{and our structure is also flexible in the choice of encoders.} ViT has proven its efficiency and performance on a variety of computer vision tasks. It has the advantage to efficiently process images of different scales and resolutions, thus capturing global information in the image and long-range dependencies between pixels. We utilize the model that is pretrained on ImageNet-21k, which contains 14 million images of 21,843 classes. \textcolor{black}{The image encoder is not specifically pretrained on remote sensing imagery as our primary objective is to validate the effectiveness of the multimodal structure rather than benchmark specific encoding approaches.} 

Specifically, the ViT model processes images by dividing them into a sequence of non-overlapping patches, where each patch is then flattened and linearly embedded into a fixed-size vector. For example, given an image $x \in \mathbb{R}^{H \times W \times C}$, it is reshaped to a sequence of 2D patches $x_p \in \mathbb{R}^{N \times (P^2 \times C)}$, where $H \times W$ is the shape of the original image, $C$ is the number of channels, $P \times P$ is the shape of each 2D patch, and $N = {{H \times W} \over P^2}$ is the number of patches, which is also the length of the hidden representations. For each patch, it is projected into latent space with a hidden vector size of $D_i$. Additionally, a special class token embedding is prepended to this sequence. As a result, each image is transformed to an embedding $E_i \in \mathbb{R}^{L_i \times D_i}$ where $L_i = N + 1$. Positional embeddings $E_{pos} \in \mathbb{R}^{L_i \times D_i}$ are added to these patch embeddings to retain spatial information. The sequence of resulting patch embeddings $E = E_i + E_{pos}$ is then passed through several transformer layers. Each transformer layer consists of multi-head self-attention mechanisms and feed-forward neural networks. We finally obtain the image representation $I \in \mathbb{R}^{L_i \times D_i}$.

For this experiment, we process the image to size $224 \times 224$ and use a patch size of $16 \times 16$. We set the hidden vector size $D_i$ to 768. Therefore, for each image, we have the image representation with a shape of $197 \times 768$.

\subsection{Text Tokenizer}
We use the text encoder from the pretrained CLIP model \cite{radford2021learning} as our text tokenizer. The CLIP model is trained on a large number of image-text pairs, so its text encoder is adept at capturing rich semantic information and can be well-aligned with visual data, \textcolor{black}{making it a practical choice for this study. However, the framework is designed to be modular, allowing for alternative text encoders to be integrated seamlessly.}

The descriptive text from VLM is first converted into token sequences, where each token represents a word or word unit. All token sequences are truncated or padded to $L_t$ tokens, including start of sequence (SOS) and end of sequence (EOS) tokens. The processed token sequence is converted into a vector with dimension $D_t$. This vector is then passed through a multilayer transformer similar to ViT to capture dependencies within the text sequence. Finally, we obtain the text representation $T \in \mathbb{R}^{L_t \times D_t}$.

For this experiment, we process the text tokens to the maximum length that the CLIP text encoder can accept, which is 77. We set the hidden vector size $D_t$ to 512. Therefore, for each description, we have the text representation with a shape of $77 \times 512$. We will also discuss the effect of fine-tuning for the pretrained encoders in Section~\ref{sec:result}.

\subsection{Multimodal Dual-Attention Encoder}
The core innovation of our methodology lies in the multimodal dual-attention encoder, which integrates textual and visual information using a series of attention mechanisms. This encoder is designed to effectively combine the strengths of both modalities, ensuring that the final representation leverages the rich semantic content from text and the detailed visual features from images. This mechanism is inspired by the Segment Anything Model (SAM) \cite{kirillov2023segment}, which employs a similar strategy for combining different types of inputs.

\begin{figure}[h]
\begin{center}
		\includegraphics[width=1.0\columnwidth]{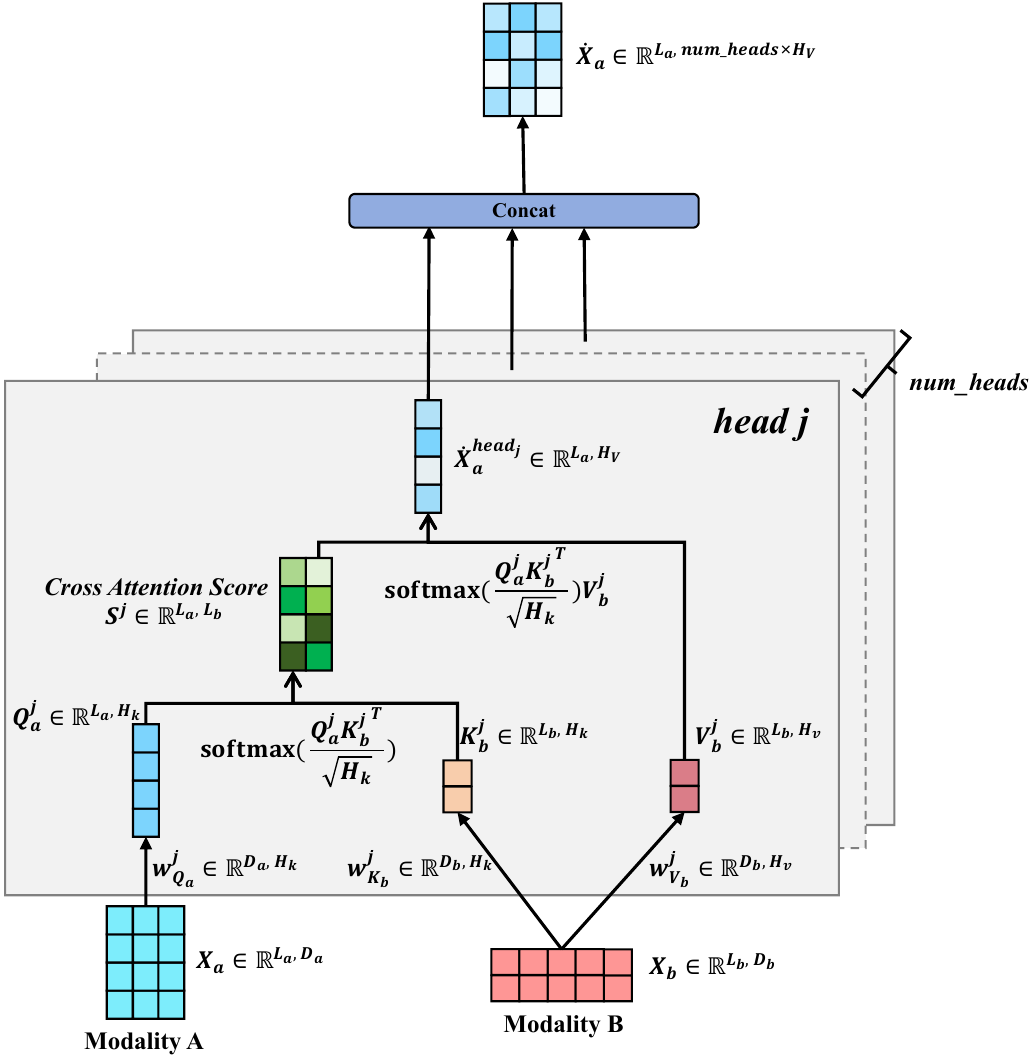}
	\caption{Cross attention mechanism illustration.}
\label{fig:cross}
\end{center}
\end{figure}

\textbf{Multi-head Cross-modal Attention Module: } Fig.~\ref{fig:cross} illustrates the computation process for the Cross-modal attention module. This cross-attention step allows representation from one modality to incorporate relevant information from the other modality, thereby enhancing the embeddings with contextually appropriate features from other modalities. As we use this module twice for image-to-text and text-to-image cross-modal attention, we denote the modalities as A and B in the figure for better understanding. Consider Modality A and Modality B, where the representations extracted from both modalities are denoted as $X_a \in \mathbb{R}^{L_a \times D_a}$ and $X_b \in \mathbb{R}^{L_b \times D_b}$, where $L$s and $D$s are the length and dimension of the representations, respectively.
The computation of the Modality A to Modality B cross-attention proceeds as follows: we define the projection for queries $Q_a^j \in \mathbb{R}^{L_a \times H_k}$, keys $K_b^j \in \mathbb{R}^{L_b \times H_k}$, and values $V_b^j \in \mathbb{R}^{L_b \times H_v}$ as:
\begin{equation}
\begin{aligned}
Q_a^j = X_a \cdot W_{Q_a}^j,   
K_b^j = X_b \cdot W_{K_b}^j,   
V_b^j = X_b \cdot W_{V_b}^j
\end{aligned}
\end{equation}
where
$W_{Q_a}^j \in \mathbb{R}^{D_a \times H_k},
W_{K_b}^j \in \mathbb{R}^{D_b \times H_k},
W_{V_b}^j \in \mathbb{R}^{D_b \times H_v}$, $H_k$ and $H_v$ are the hidden vector dimensions for $K_b^j$ and $V_b^j$ respectively, $j$ is the head number, and typically $H_v = {D_a \over num_heads}$.
The cross-modal attention is denoted as:
\begin{equation}
\begin{aligned}
\dot{X_a^{head_j}} &= {\operatorname{CMA} (Q_a^j, K_b^j, V_b^j)}\\ &= \operatorname{softmax}\left(\frac{Q_a^j \cdot {K_b^j\top}}{\sqrt{H_k}}\right)\cdot{V_b^j}
\end{aligned}
\end{equation}
then $\dot{X_a^{head_j}} \in \mathbb{R}^{L_a \times H_v}$ are concatenated for all heads:
\begin{equation}
\begin{aligned}
\dot{X_a}  \in \mathbb{R}^{L_a \times D_a} &= {\operatorname{Concat} (\dot{X_a^{0}}, \dot{X_a^{1}}, \cdots , \dot{X_a}^{num\_heads}})
\end{aligned}
\end{equation}

\textbf{Multimodal Dual-Attention Encoder}: The process begins with self-attention for the input text representation as is shown in Fig.~\ref{fig:overview}. This step is crucial as it refines the contextual embeddings within the text representations, ensuring that dependencies and relationships across the entire sequence are captured. Mathematically, the self-attention operation can be expressed as: 
\begin{equation}
\begin{aligned}
X_t &= \operatorname{SelfAttention}(Q_t,K_t,V_t) \\
&= \operatorname{softmax}({Q_t{K_t}^T \over \sqrt{d_k}}V_t ) 
\end{aligned}
\end{equation}
where $Q_t,K_t$, and $V_t$are projections of the text representation, $d_k^i$ is the vector dimension for $K_t^i$.
Following self-attention, the refined text representation is used as the query in the cross-attention mechanism, where it interacts with the image representation, which acts as the key and value to perform text-to-image cross-attention. The enhanced text representation $\dot{X_t}$ is then added to $X_t$, and a layer normalization is applied to form the final text representation $\hat{X_t}$:
\begin{equation}
\begin{aligned}
\hat{X_t} &= \operatorname{LayerNorm}(\dot{X_t} + X_t) \\
\end{aligned}
\end{equation}
The enhanced text representation $\dot{X_t}$ is first put into a Multilayer Perceptron (MLP), and the output is used as the reference to calculate cross-attention with image representation. We finally get the enhanced image representation $\dot{X_i}$ (the output of image-to-text cross attention is first projected back to the same  dimension with the image representation $I$). We add $\dot{X_i}$ to $X_i$ and apply a layer normalization:
\begin{equation}
\begin{aligned}
\hat{X_i} &= \operatorname{LayerNorm}(\dot{X_i} + X_i) \\
\end{aligned}
\end{equation}
Note that the Multimodal Dual-Attention Encoder consists of $N$ identical layers, where the output of layer $(n-1)$ serves as the input to layer $n$.

\textbf{Prediction Layer}:
The final representations from the last layer of the Multimodal Dual-Attention Encoder, denoted as $\hat{X_i}$ and $\hat{X_t}$, are concatenated and passed through linear layers followed by softmax functions to generate the scene type prediction. For model optimization, we utilize the multi-class cross-entropy loss function.

\begin{figure*}[h]
\begin{center}
		\includegraphics[width=1.0\textwidth]{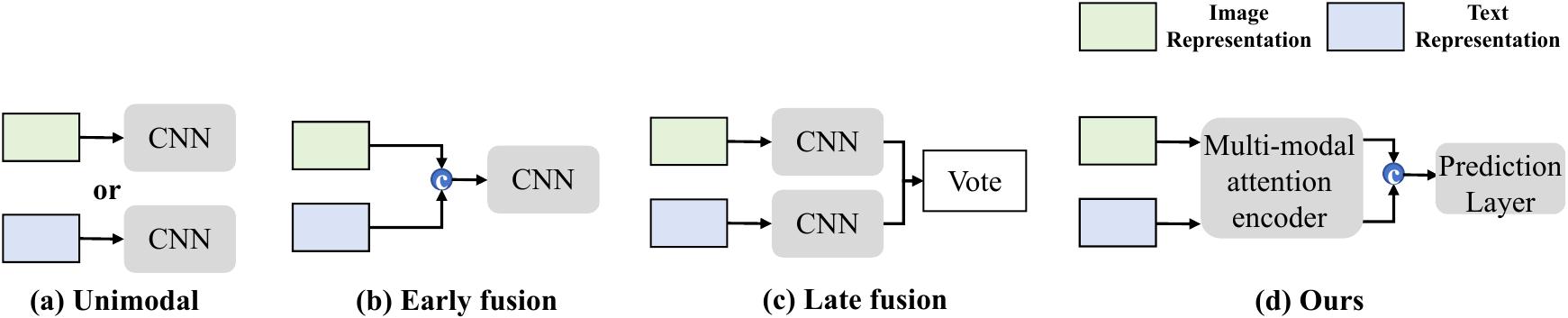}
	\caption{Comparison of Baseline models and our model. (a) Image embeddings or text embeddings are put into the CNN model for unimodal training; (b) Image and text embeddings are first concatenated and then put into the CNN model; (c) Image embeddings and text embeddings are put into two CNN models separately, the decisions are fused by voting; (d) Our multimodal dual-attention network.}
\label{fig:baseline}
\end{center}
\end{figure*}

\section{Experiments}\label{sec:EXPERIMENTS}
In this section, we will detail the experiments designed to evaluate the effectiveness of our proposed model. Our primary aim is to enhance remote sensing scene classification by combining VLM-generated text descriptions with image data. Specifically, we seek to address the following research questions: 
\begin{enumerate}
    \item Can we improve remote sensing scene classification accuracy by integrating VLM-generated text descriptions with image data?
    \item Are VLM-generated text descriptions more effective than human-annotated and automatically annotated text descriptions in improving classification performance?
    \item Can a multimodal dual-attention mechanism better address the complementary information between modalities compared to traditional fusion strategies?
\end{enumerate}

To this end, we compared our model with baseline models utilizing both unimodal and various multimodal fusion strategies, including early fusion and late fusion approaches. We conducted extensive evaluations of our model across a wide range of remote sensing datasets to comprehensively assess its performance. We also evaluate our models' performance on an image captioning dataset to compare the effectiveness of VLM-generated descriptions against human-annotated descriptions.

\subsection{Dataset Decription}\label{sec：dataset}
Experiments were conducted on five scene classification datasets for quantitative and qualitative analysis. Additionally, a manually annotated image captioning dataset was used to evaluate the validity of the VLM-generated descriptions.

\textbf{Aerial Image Dataset (AID):} A large-scale aerial image dataset containing 10,000 images across 30 classes. All images are labeled by experts in the field of remote sensing image interpretation. Each image has a size of 600$\times$600 pixels with resolutions of up to half a meter.

\textbf{PatternNet:} A large-scale dataset containing 38 classes of high-resolution aerial images. Each class has 800 images of 256$\times$256 pixels, with spatial resolutions ranging from 0.062 to 4.693 meters.

\textbf{Million-AID:} The state-of-the-art largest scene classification dataset, containing 51 classes with over one million aerial images at various spatial resolutions ranging from 0.5 to 153 meters. It features a hierarchical structure with 8 main categories, 28 second-level nodes, and 51 leaf categories. For this study, we randomly sampled 10,000 images from the dataset. We established two experimental settings: (1) Million-AID Level 3: Utilizing all 51 leaf categories; (2) Million-AID Level 2: Comprising 38 classes (28 second-level nodes plus 10 leaf categories from branches without intermediate nodes).

\textbf{DeepGlobe:} A challenge dataset for land cover classification from satellite images, introduced at the CVPR 2018 DeepGlobe workshop. For our experiment, we processed the dataset as follows: first, we split the original images into non-overlapping 600$\times$600 pixel patches. Next, we selected patches where over 75\% of the area belongs to a single land cover class and assigned each selected patch the label of the dominant class. This processing resulted in a dataset of 9,654 images across 6 land cover classes. We randomly sampled a balanced dataset of 3,000 images.

\textbf{UCM-captioning:} This dataset was adapted in 2017 for remote sensing (RS) image captioning and retrieval tasks. It comprises 2,100 high-resolution aerial images across 21 land cover classes, with 100 images per class. Each image is 256$\times$256 pixels in size and is accompanied by 5 human-annotated text descriptions, providing similar content but with varying descriptive styles. For our experiment, we randomly selected one of the five descriptions for each image to serve as the text modality. This dataset was used as a benchmark to validate the effectiveness of our VLM-generated descriptions, allowing us to compare the performance of our model using automatically generated text against human-annotated captions for remote sensing scene classification tasks.

All datasets were partitioned into training, validation, and test sets using a consistent methodology. First, 10\% of the entire dataset was reserved for the test set. The remaining 90\% was then further divided into training and validation sets with a ratio of 8:2, respectively. This results in an overall split ratio of approximately 72\% for training, 18\% for validation, and 10\% for testing. We also investigate other train-validation splitting ratios in later sections.

\subsection{Baseline and Ablation models}

To evaluate the effectiveness of our multimodal approach, we implemented several baseline models including two unimodal approaches and two fusion approaches as shown in Fig.~\ref{fig:baseline} (a), (b) and (c):

\textbf{Image-Only (\textit{IO}) Model:} We employed a Convolutional Neural Network (CNN) to classify representations derived solely from the ViT encoder. The CNN architecture consists of two 2D convolutional layers. The first layer uses 128 filters with a kernel size of 3 and stride of 1, followed by ReLU activation and max-pooling (pool size 2). The second convolutional layer uses 256 filters with the same kernel size and stride, again followed by ReLU activation and max-pooling. The output is then flattened and passed through two fully connected layers with 1024 and 512 units respectively, both with ReLU activation. After that, the learned representation is put to the final softmax layer for multi-class classification.

\textbf{Text-Only (\textit{TO}) Model:} Text representations from the CLIP text encoder are put in a similar CNN architecture with the image-only model, with adjustments for the input shape. It consists of two 1D convolutional layers: the first with 64 filters and the second with 128 filters, both using a kernel size of 3 and stride of 1.

\textbf{Early Fusion (\textit{EF}) Model:} We first transform the image representation from its original shape of (197, 768) to (197, 512) using a linear projection layer to align with the dimensionality of the text representations. The transformed image representation is then concatenated with the text representation along the feature dimension, resulting in a combined input of shape (274, 512). This fused representation is then fed into a CNN similar to the unimodal models for classification.

\begin{table*}[h]
\centering
\setlength{\tabcolsep}{5pt}
\renewcommand{\arraystretch}{1.5}
\caption{Summary of experimental results in terms of average and standard deviations of Overall Accuracy (OA\%), Average Accuracy (AA\%), and \textit{Kappa} coefficient (\textit{Kappa\%}).}
\label{table:pp}
 \resizebox{1.0\textwidth}{!}{
\begin{tabular}{c|ccc|ccc|ccc|ccc|ccc}
\hline
Dataset   & \multicolumn{3}{c|}{AID}                                                                                                                                                     & \multicolumn{3}{c|}{PatternNet}                                                                                                                                              & \multicolumn{3}{c|}{Million-AID-2}                                                                                                                                          & \multicolumn{3}{c|}{Million-AID-3}                                                                                                                                          & \multicolumn{3}{c}{DeepGlobe}                                                                                                                                               \\
Baselines & OA/top-5                                                       & AA                                                   & \textit{Kappa}                                                & OA/top-5                                                       & AA                                                   & \textit{Kappa}                                                & OA/top-5                                                      & AA                                                   & \textit{Kappa}                                                & OA/top-5                                                      & AA                                                   & \textit{Kappa}                                                & OA/top-3                                                      & AA                                                   & \textit{Kappa}                                                \\ \hline \hline
\textit{IO}        & \begin{tabular}[c]{@{}c@{}}95.6/99.8\\ (0.4/0.1)\end{tabular}  & \begin{tabular}[c]{@{}c@{}}94.3\\ (0.5)\end{tabular} & \begin{tabular}[c]{@{}c@{}}94.8\\ (0.3)\end{tabular} & \begin{tabular}[c]{@{}c@{}}98.0/99.8\\ (0.5/0.1)\end{tabular}  & \begin{tabular}[c]{@{}c@{}}97.5\\ (0.4)\end{tabular} & \begin{tabular}[c]{@{}c@{}}97.0\\ (0.3)\end{tabular} & \begin{tabular}[c]{@{}c@{}}88.3/96.1\\ (0.2/0.2)\end{tabular} & \begin{tabular}[c]{@{}c@{}}87.8\\ (0.3)\end{tabular} & \begin{tabular}[c]{@{}c@{}}87.3\\ (0.3)\end{tabular} & \begin{tabular}[c]{@{}c@{}}85.9/96.9\\ (0.4/0.2)\end{tabular} & \begin{tabular}[c]{@{}c@{}}85.5\\ (0.4)\end{tabular} & \begin{tabular}[c]{@{}c@{}}85.0\\ (0.4)\end{tabular} & \begin{tabular}[c]{@{}c@{}}69.9/85.2\\ (0.5/0.3)\end{tabular} & \begin{tabular}[c]{@{}c@{}}69.0\\ (0.5)\end{tabular} & \begin{tabular}[c]{@{}c@{}}68.4\\ (0.4)\end{tabular} \\ \hline
\textit{TO}        & \begin{tabular}[c]{@{}c@{}}89.5/97.0\\ (1.1/0.1)\end{tabular}  & \begin{tabular}[c]{@{}c@{}}88.0\\ (1.0)\end{tabular} & \begin{tabular}[c]{@{}c@{}}88.5\\ (0.9)\end{tabular} & \begin{tabular}[c]{@{}c@{}}92.1/97.3\\ (0.8/0.2)\end{tabular}  & \begin{tabular}[c]{@{}c@{}}91.5\\ (0.9)\end{tabular} & \begin{tabular}[c]{@{}c@{}}91.0\\ (0.7)\end{tabular} & \begin{tabular}[c]{@{}c@{}}84.5/95.5\\ (0.4/0.3)\end{tabular} & \begin{tabular}[c]{@{}c@{}}83.4\\ (0.4)\end{tabular} & \begin{tabular}[c]{@{}c@{}}83.6\\ (0.4)\end{tabular} & \begin{tabular}[c]{@{}c@{}}80.6/95.2\\ (0.5/0.4)\end{tabular} & \begin{tabular}[c]{@{}c@{}}78.5\\ (0.4)\end{tabular} & \begin{tabular}[c]{@{}c@{}}79.0\\ (0.4)\end{tabular} & \begin{tabular}[c]{@{}c@{}}63.3/89.7\\ (0.9/0.8)\end{tabular} & \begin{tabular}[c]{@{}c@{}}62.5\\ (1.0)\end{tabular} & \begin{tabular}[c]{@{}c@{}}62.0\\ (0.8)\end{tabular} \\ \hline
\textit{EF}        & \begin{tabular}[c]{@{}c@{}}95.5/99.8\\ (0.8/0.1)\end{tabular}  & \begin{tabular}[c]{@{}c@{}}94.0\\ (0.7)\end{tabular} & \begin{tabular}[c]{@{}c@{}}94.5\\ (0.6)\end{tabular} & \begin{tabular}[c]{@{}c@{}}97.2/99.7\\ (0.5/0.1)\end{tabular}  & \begin{tabular}[c]{@{}c@{}}96.5\\ (0.5)\end{tabular} & \begin{tabular}[c]{@{}c@{}}96.0\\ (0.5)\end{tabular} & \begin{tabular}[c]{@{}c@{}}88.5/96.5\\ (0.5/0.4)\end{tabular} & \begin{tabular}[c]{@{}c@{}}88.2\\ (0.5)\end{tabular} & \begin{tabular}[c]{@{}c@{}}88.4\\ (0.5)\end{tabular} & \begin{tabular}[c]{@{}c@{}}85.6/97.6\\ (0.4/0.4)\end{tabular} & \begin{tabular}[c]{@{}c@{}}83.5\\ (0.4)\end{tabular} & \begin{tabular}[c]{@{}c@{}}85.1\\ (0.3)\end{tabular} & \begin{tabular}[c]{@{}c@{}}79.7/94.2\\ (0.3/0.3)\end{tabular} & \begin{tabular}[c]{@{}c@{}}79.0\\ (0.4)\end{tabular} & \begin{tabular}[c]{@{}c@{}}78.5\\ (0.3)\end{tabular} \\ \hline
\textit{LF}        & \begin{tabular}[c]{@{}c@{}}95.9/99.8\\ (0.7/0.1)\end{tabular}  & \begin{tabular}[c]{@{}c@{}}94.5\\ (0.6)\end{tabular} & \begin{tabular}[c]{@{}c@{}}94.9\\ (0.5)\end{tabular} & \begin{tabular}[c]{@{}c@{}}97.8/99.6\\ (0.5/0.0)\end{tabular}  & \begin{tabular}[c]{@{}c@{}}97.0\\ (0.6)\end{tabular} & \begin{tabular}[c]{@{}c@{}}96.5\\ (0.4)\end{tabular} & \begin{tabular}[c]{@{}c@{}}88.3/97.0\\ (0.7/0.4)\end{tabular} & \begin{tabular}[c]{@{}c@{}}87.9\\ (0.5)\end{tabular} & \begin{tabular}[c]{@{}c@{}}88.0\\ (0.5)\end{tabular} & \begin{tabular}[c]{@{}c@{}}85.2/96.8\\ (0.7/0.5)\end{tabular} & \begin{tabular}[c]{@{}c@{}}83.8\\ (0.3)\end{tabular} & \begin{tabular}[c]{@{}c@{}}85.2\\ (0.3)\end{tabular} & \begin{tabular}[c]{@{}c@{}}79.5/94.0\\ (0.2/0.2)\end{tabular} & \begin{tabular}[c]{@{}c@{}}79.5\\ (0.2)\end{tabular} & \begin{tabular}[c]{@{}c@{}}78.2\\ (0.3)\end{tabular} \\ \hline \hline
\textit{no CAtt}   & \begin{tabular}[c]{@{}c@{}}97.0/99.5\\ (0.7/0.1)\end{tabular}  & \begin{tabular}[c]{@{}c@{}}96.0\\ (0.6)\end{tabular} & \begin{tabular}[c]{@{}c@{}}96.5\\ (0.5)\end{tabular} & \begin{tabular}[c]{@{}c@{}}98.5/99.8\\ (0.2/0.1)\end{tabular}  & \begin{tabular}[c]{@{}c@{}}98.5\\ (0.2)\end{tabular} & \begin{tabular}[c]{@{}c@{}}98.4\\ (0.3)\end{tabular} & \begin{tabular}[c]{@{}c@{}}92.5/99.2\\ (0.8/0.6)\end{tabular} & \begin{tabular}[c]{@{}c@{}}92.0\\ (0.5)\end{tabular} & \begin{tabular}[c]{@{}c@{}}92.5\\ (0.5)\end{tabular} & \begin{tabular}[c]{@{}c@{}}90.2/97.8\\ (0.4/0.3)\end{tabular} & \begin{tabular}[c]{@{}c@{}}88.5\\ (0.3)\end{tabular} & \begin{tabular}[c]{@{}c@{}}88.9\\ (0.4)\end{tabular} & \begin{tabular}[c]{@{}c@{}}88.3/99.7\\ (0.4/0.2)\end{tabular} & \begin{tabular}[c]{@{}c@{}}88.3\\ (0.4)\end{tabular} & \begin{tabular}[c]{@{}c@{}}88.0\\ (0.2)\end{tabular} \\ \hline
\textit{ICAtt}     & \begin{tabular}[c]{@{}c@{}}97.5/99.8\\ (0.4/0.1)\end{tabular}  & \begin{tabular}[c]{@{}c@{}}96.5\\ (0.3)\end{tabular} & \begin{tabular}[c]{@{}c@{}}97.0\\ (0.2)\end{tabular} & \begin{tabular}[c]{@{}c@{}}99.2/100.0\\ (0.2/0.1)\end{tabular} & \begin{tabular}[c]{@{}c@{}}99.2\\ (0.3)\end{tabular} & \begin{tabular}[c]{@{}c@{}}99.1\\ (0.2)\end{tabular} & \begin{tabular}[c]{@{}c@{}}94.6/99.5\\ (0.8/0.4)\end{tabular} & \begin{tabular}[c]{@{}c@{}}94.1\\ (0.7)\end{tabular} & \begin{tabular}[c]{@{}c@{}}94.3\\ (0.6)\end{tabular} & \begin{tabular}[c]{@{}c@{}}92.5/99.5\\ (0.5/0.4)\end{tabular} & \begin{tabular}[c]{@{}c@{}}91.6\\ (0.3)\end{tabular} & \begin{tabular}[c]{@{}c@{}}92.0\\ (0.3)\end{tabular} & \begin{tabular}[c]{@{}c@{}}85.3/99.7\\ (0.2/0.1)\end{tabular} & \begin{tabular}[c]{@{}c@{}}85.3\\ (0.2)\end{tabular} & \begin{tabular}[c]{@{}c@{}}84.8\\ (0.3)\end{tabular} \\ \hline
\textit{TCAtt}     & \begin{tabular}[c]{@{}c@{}}97.2/99.9\\ (0.6/0.0)\end{tabular}  & \begin{tabular}[c]{@{}c@{}}96.2\\ (0.5)\end{tabular} & \begin{tabular}[c]{@{}c@{}}96.7\\ (0.4)\end{tabular} & \begin{tabular}[c]{@{}c@{}}99.0/100.0\\ (0.5/0.1)\end{tabular} & \begin{tabular}[c]{@{}c@{}}99.0\\ (0.5)\end{tabular} & \begin{tabular}[c]{@{}c@{}}98.9\\ (0.4)\end{tabular} & \begin{tabular}[c]{@{}c@{}}93.7/99.5\\ (0.6/0.5)\end{tabular} & \begin{tabular}[c]{@{}c@{}}93.2\\ (0.6)\end{tabular} & \begin{tabular}[c]{@{}c@{}}93.5\\ (0.6)\end{tabular} & \begin{tabular}[c]{@{}c@{}}92.0/99.6\\ (0.5/0.4)\end{tabular} & \begin{tabular}[c]{@{}c@{}}91.0\\ (0.3)\end{tabular} & \begin{tabular}[c]{@{}c@{}}92.0\\ (0.3)\end{tabular} & \begin{tabular}[c]{@{}c@{}}89.7/99.9\\ (0.5/0.2)\end{tabular} & \begin{tabular}[c]{@{}c@{}}89.7\\ (0.5)\end{tabular} & \begin{tabular}[c]{@{}c@{}}87.6\\ (0.4)\end{tabular} \\ \hline
\textbf{Ours}      & \begin{tabular}[c]{@{}c@{}}\textbf{98.9/100.0}\\ (0.8/0.0)\end{tabular} & \begin{tabular}[c]{@{}c@{}}\textbf{98.1}\\ (0.3)\end{tabular} & \begin{tabular}[c]{@{}c@{}}\textbf{97.0}\\ (0.2)\end{tabular} & \begin{tabular}[c]{@{}c@{}}\textbf{99.4/100.0}\\ (0.3/0.0)\end{tabular} & \begin{tabular}[c]{@{}c@{}}\textbf{99.4}\\ (0.4)\end{tabular} & \begin{tabular}[c]{@{}c@{}}\textbf{98.5}\\ (0.3)\end{tabular} & \begin{tabular}[c]{@{}c@{}}\textbf{97.4/99.4}\\ (0.5/0.1)\end{tabular} & \begin{tabular}[c]{@{}c@{}}\textbf{97.1}\\ (0.5)\end{tabular} & \begin{tabular}[c]{@{}c@{}}\textbf{97.0}\\ (0.4)\end{tabular} & \begin{tabular}[c]{@{}c@{}}\textbf{95.6/99.8}\\ (0.8/0.2)\end{tabular} & \begin{tabular}[c]{@{}c@{}}\textbf{95.0}\\ (0.7)\end{tabular} & \begin{tabular}[c]{@{}c@{}}\textbf{94.5}\\ (0.6)\end{tabular} & \begin{tabular}[c]{@{}c@{}}\textbf{91.3/99.9}\\ (1.2/0.5)\end{tabular} & \begin{tabular}[c]{@{}c@{}}\textbf{90.5}\\ (1.1)\end{tabular} & \begin{tabular}[c]{@{}c@{}}\textbf{90.0}\\ (1.0)\end{tabular} \\ \hline
\end{tabular}}
\end{table*}

\textbf{Late Fusion (\textit{LF}) Model:} We employ separate CNNs to process the image and text representations independently before combining their outputs. The outputs of these two branches are then weighted and combined. Specifically, we use learnable scalar weights $\alpha$ and $\beta$ for the image and text features respectively, where $\alpha$ and $\beta$ were initialized to 0.5 each and were updated during training. The fused representation $f_{fusion}$ is computed as: $f_{fusion} = \alpha f_{image} + \beta f_{text}$. 

To better understand the function of different components in our proposed model, we conducted several ablation studies by removing certain modules:

\textbf{No Cross Attention(\textit{no CAtt}) Model:} We remove the multi-head cross-modal attention module for both modalities. The image-to-text cross-attention module is replaced with a self-attention layer, while the text-to-image cross-attention module and the subsequent MLP layer are entirely removed. Consequently, there is no information exchange between the two modalities prior to their concatenation. 

\textbf{Image-to-text Cross Attention Only (\textit{ICAtt}) Model:} We remove the text-to-image cross-attention module. Consequently, the text representation after the self-attention module is not enhanced by information from the image representation. 

\textbf{Text-to-Image Cross Attention Only (\textit{TCAtt}) Model:} We substitute the image-to-text cross-attention module with a self-attention layer. Additionally, the MLP layer following the text-to-image cross-attention module is removed. As a result, information flows only from the image representation to the text representation.

\subsection{Evaluation Scheme and Metrics} 
To mitigate potential biases introduced by a single data split, after setting aside 10\% of the data as a held-out test set, we randomly shuffled the remaining data and divided it into 5 equal folds where we implemented a 5-fold cross-validation strategy. 
We employ multiple evaluation metrics to comprehensively assess our model's performance, including overall accuracy (\textit{OA}), average accuracy (\textit{AA}), and the \textit{Kappa} coefficient (\textit{Kappa}). Specifically, overall accuracy measures the proportion of correctly classified samples across all classes, while average accuracy calculates the mean of the individual class accuracies, and \textit{Kappa} coefficient evaluates the model's reliability over imbalance datasets. We also calculated the top-5 overall accuracy for most datasets, and the top-3 accuracy specifically for the DeepGlobe dataset (due to its limited number of six classes). Additionally, we utilize confusion matrices to provide a detailed breakdown of the model's classification results.

\subsection{Implementation Details}
We conducted our experiments on an NVIDIA GeForce RTX 3090 GPU. The model was implemented with a multi-head attention mechanism, utilizing 4 heads for both self and cross-attention layers, with an attention dropout rate of 0.05. The multimodal dual-attention encoder consisted of 2 layers. We applied dropout rates of 0.1 to both the MLP ReLU layers and the residual blocks to prevent overfitting. The model was trained using the Adam optimizer with an initial learning rate of 1e-4, which was decayed after 20 epochs. We employed a batch size of 64 and trained the model for a total of 40 epochs. To stabilize training, we implemented gradient clipping with a value of 0.8. Early stopping was employed with a patience of 10 epochs to prevent overfitting and optimize training time.

\section{Results and Analysis}
\label{sec:result}
\subsection{Quantitative Measurements}

The experimental results presented in Table~\ref{table:pp} and Table~\ref{table:ucm} provide comprehensive evidence of our proposed method's performance. Table~\ref{table:pp} offers a detailed comparison with baseline models and ablation variants across five diverse datasets while Table~\ref{table:ucm} offers an evaluation of human-annotated captions versus VLM-generated descriptions based on the UCM-captioning dataset. The results convincingly demonstrate that our model achieves superior performance compared to other baselines and ablation models, effectively addressing our primary research questions. A detailed analysis of these comparisons is summarized as follows:

\textbf{Comparison with baselines:} By comparing the results of Image-Only (\textit{IO}), Text-Only (\textit{TO}), Early Fusion (\textit{EF}), and Late Fusion (\textit{LF}) models with our proposed model, we observe that our model consistently achieves the best performance across all experimental settings. Notably, the \textit{IO} setting consistently outperforms the \textit{TO} setting by a margin of 3.8\% to 6.6\% across various datasets. This discrepancy can be attributed to several factors. Firstly, VLM-generated textual descriptions may occasionally be inaccurate or incomplete, particularly when dealing with low-resolution images or scenes with unclear elements, posing challenges for the VLM in accurate interpretation. Secondly, the process of converting complex visual information into concise textual descriptions inevitably results in some loss of detail and specificity crucial for precise classification, especially when the VLM used is not pretrained with domain-specific knowledge.

When comparing \textit{IO} and \textit{TO} with \textit{EF} and \textit{LF} models, we observe that in most cases, early fusion does not outperform the image-only approach, while late fusion achieves the best results among these four methods. This indicates that multimodal fusion can be beneficial only when the modalities are well integrated. The underperformance of \textit{EF} may be attributed to the fact that concatenated features from image and text modalities might not be effectively integrated in the early stages of the network, leading to suboptimal feature representations. In contrast, \textit{LF} employs separate networks to process image and text representations independently, allowing each modality to learn rich and detailed features specific to its data type without interference from the other modality early in the process.

Ultimately, our proposed model outperforms the best of these four baseline approaches by a margin ranging from 1.4\% to 11.6\%, demonstrating the effectiveness of our multimodal fusion strategy for RSSC tasks. This superior performance can be attributed to our model's dual-attention mechanism, which considers both image-to-text and text-to-image cross-modal relationships. Unlike simpler early or late fusion techniques, our approach enables a more nuanced integration of the two modalities, allowing the model to capture complex, bidirectional interactions between visual and textual features. The comparison with baseline models provides a compelling answer to our first research question.

\textbf{Comparison with ablation models:} By comparing the performance of the No Cross-Attention (\textit{No CAtt}), Image-to-Text Cross-Attention (\textit{ICAtt}), and Text-to-Image Cross-Attention (\textit{TCAtt}) models with our full model, we can demonstrate the effectiveness of the dual-attention mechanism.

First, when comparing \textit{No CAtt} with previous fusion methods, we observe an improvement of 0.5\% to 8.6\% over the best baseline model, indicating that the Transformer-based approach outperforms CNN models in extracting effective representations. This superiority can be attributed to the Transformer's ability to capture long-range dependencies and contextual relationships, which are crucial for RSSC tasks where multiple objects often coexist within a single image.

Next, comparing \textit{No CAtt} with \textit{ICAtt} and \textit{TCAtt} models, we observe that \textit{No CAtt} consistently underperforms, underscoring that both directions of cross-modal attention significantly enhance the model's effectiveness. Interestingly, for most datasets, the \textit{ICAtt} model outperforms the \textit{TCAtt} model by a margin of 0.2\% to 0.9\%. This suggests that image representations benefit more from cross-attention enhancement from the text modality, as textual descriptions provide focused, semantic information that guides the model’s attention to relevant visual features in the image.

Finally, our proposed model effectively leverages the dual-attention mechanism, consistently extracting and refining the most salient features from both modalities for RSSC tasks. As a result, our full model achieves a significant improvement of 0.2\% to 3.1\% compared to the best-performing ablation models.

The comparison with ablation models provides a clear answer to our third research question.
\begin{table}[]
\centering
\renewcommand{\arraystretch}{1.5}
\caption{Comparison of experimental results on UCM-Caption dataset with human-annotated captions and VLM captions in terms of Overall Accuracy (OA\%).}
\label{table:ucm}
 \resizebox{0.4\textwidth}{!}{
\begin{tabular}{c|c|c|c}
\hline
Metrics                  & OA   & AA   & Kappa \\ \hline \hline
Human-annotated captions & 96.4 & 96.4 & 96.2  \\ \hline
VLM captions             & 99.3 & 99.3 & 99.0  \\ \hline
\end{tabular}
}
\end{table}

\begin{figure*}[!htb]
\begin{center}
		\includegraphics[width=1\textwidth]{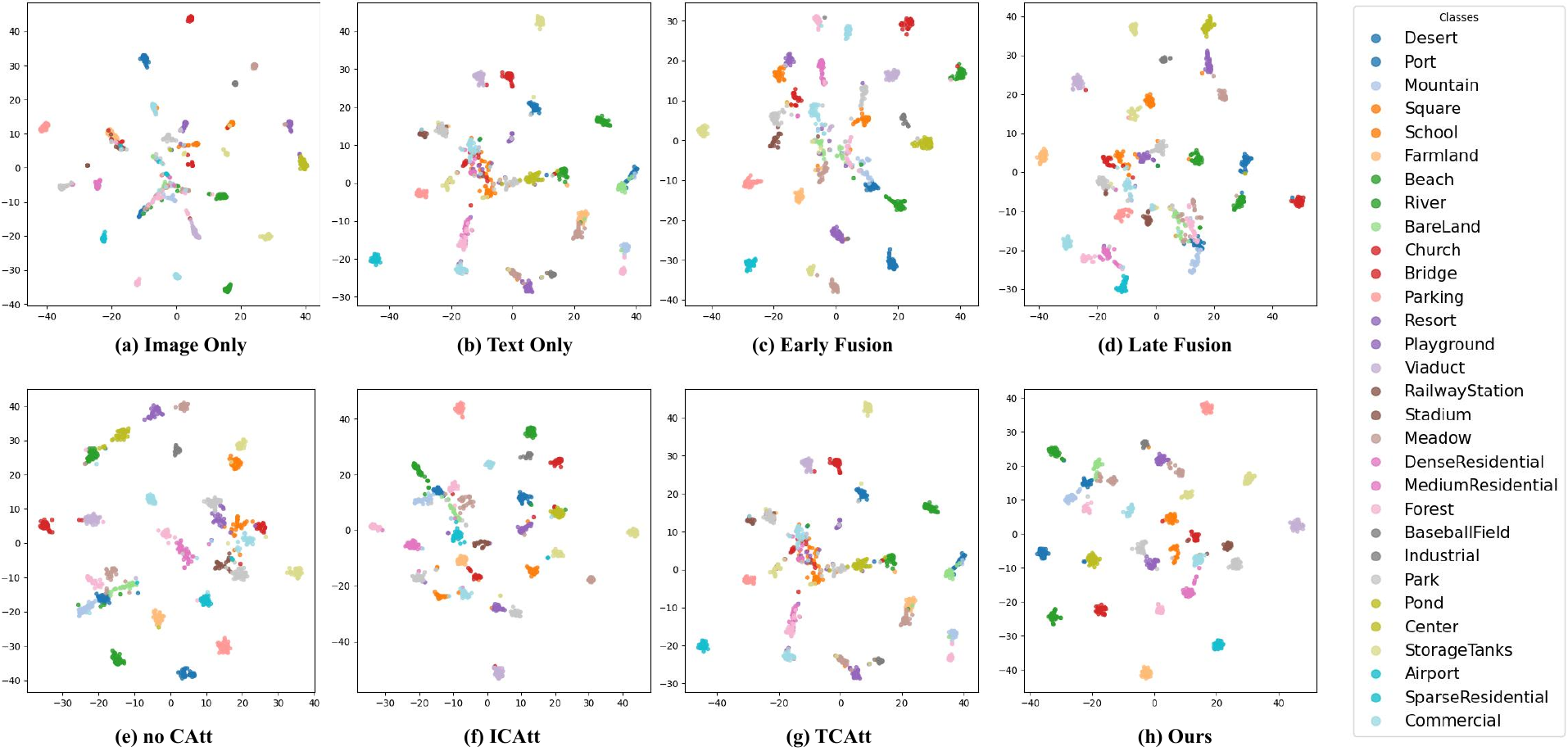}
	\caption{TSNE visualization of the AID test set. }
\label{fig:tsne}
\end{center}
\end{figure*}

\begin{figure*}[!htb]
\begin{center}

		\includegraphics[width=1.0\textwidth]{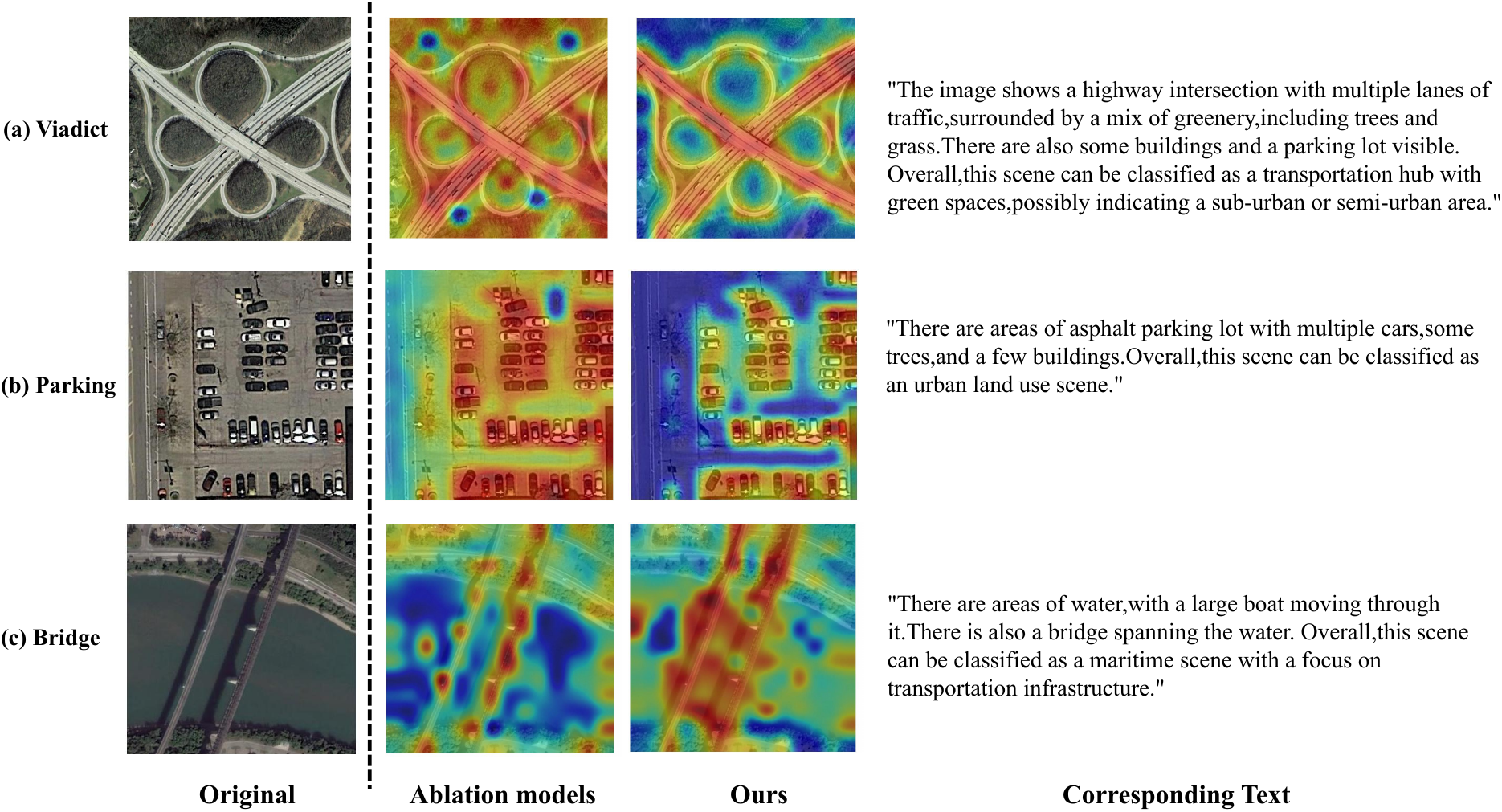}
	\caption{Attention Heatmap Comparison Across Models for AID Test Set. Attention heatmap for ablation models from top to bottom: No Cross-Attention (\textit{No CAtt}), Image-to-Text Cross-Attention (\textit{ICAtt}), and Text-to-Image Cross-Attention (\textit{TCAtt}) models.}
\label{fig:heat}
\end{center}
\end{figure*}

\begin{figure*}[!htb]
\begin{center}

		\includegraphics[width=\textwidth]{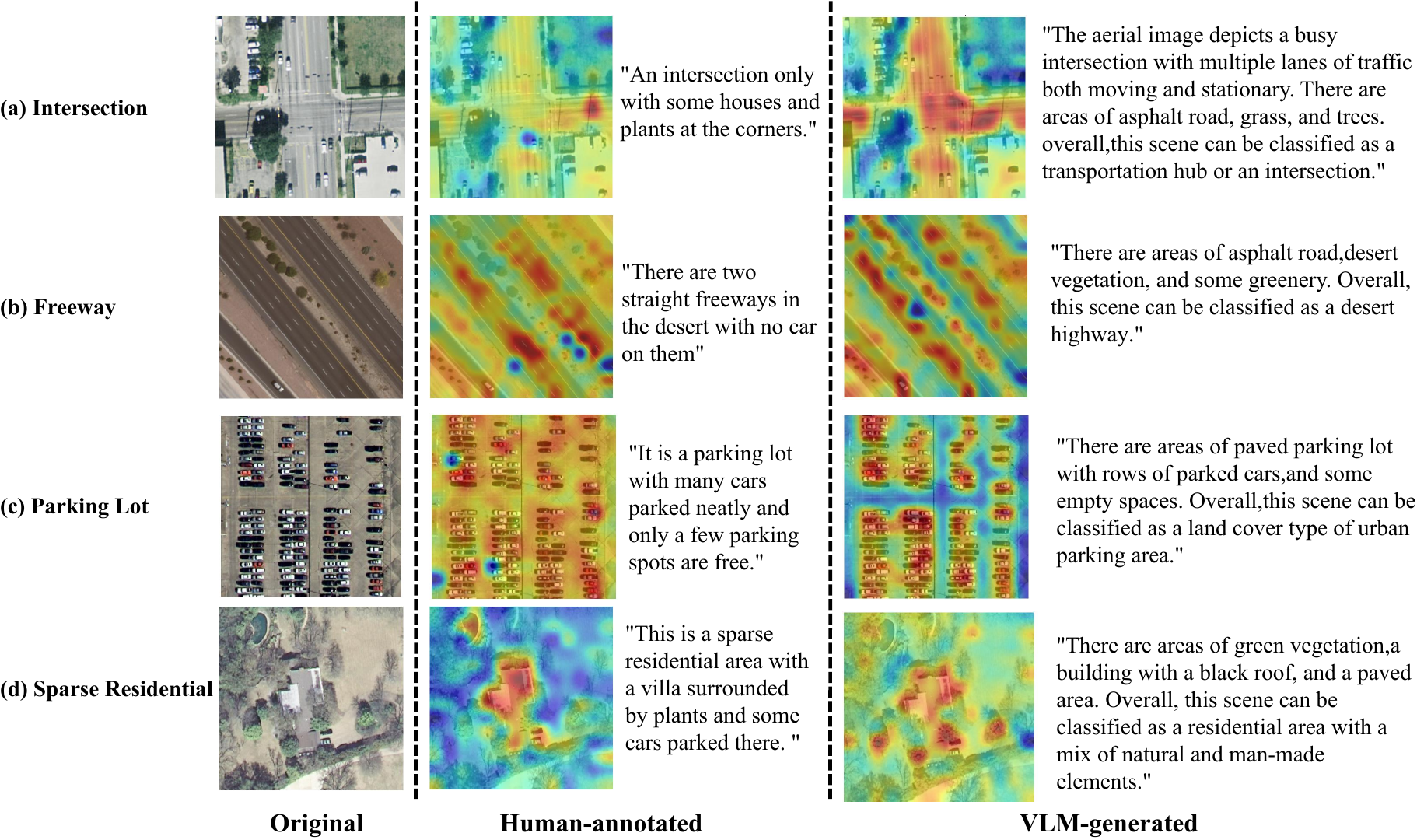}
	\caption{Comparison of heatmaps of the UCM-caption test set with human-annotated caption and VLM captions. }
\label{fig:ucm}
\end{center}
\end{figure*}

\textbf{Comparison of different text sources:} To further evaluate our model's performance and assess the efficacy of VLM-generated descriptions, we conducted additional experiments using the UCM-captioning dataset. In these experiments, we compared the performance of our model using two different text modalities while keeping image modality the same: human-annotated captions (selecting the longest of the five available captions for each image) and VLM-generated descriptions. Examples of both human-annotated captions and VLM-generated descriptions are illustrated in Fig.~\ref{fig:ucm}. 

The experimental results in Table~\ref{table:ucm}  reveal that our model utilizing VLM-generated descriptions outperforms the version using human-annotated captions by a margin of 2.9\%. This improvement suggests that VLM-generated descriptions can provide more effective textual information than simple human annotations. This finding addresses our second research question and highlights the potential of VLMs in enhancing remote sensing scene classification tasks.

\subsection{Visualizations}

\textbf{T-SNE Visualization Comparison: } Fig.\ref{fig:tsne}(a)-(h) presents t-SNE visualizations of the feature representations before the final projection layer for each model, using the AID test set. Overall, we observe that the Image-Only t-SNE (Fig.\ref{fig:tsne}a) displays more compact clusters compared to the Text-Only t-SNE (Fig.~\ref{fig:tsne}b), \textcolor{black}{e.g. the comparison of pink circle and blue circle clusters in both tsne figures,} indicating that image features tend to form tighter groupings. However, both modalities exhibit instances of overlapping clusters, though for different class pairs. For instance, the Port and Parking classes overlap in the Image-Only representation but are distinctly separated in the Text-Only visualization, demonstrating how both image and text modalities contribute unique information for class distinction.

When comparing our proposed model's t-SNE plot (Fig.~\ref{fig:tsne}h) with those of baseline and ablation models, we observe a significant improvement in class separation. Our model's visualization exhibits more distinct and well-defined clusters, with larger inter-class distances and more compact intra-class groupings, indicating enhanced feature representations and improved inter-class discrimination. In contrast, the baseline models' t-SNE plots show more scattered clusters, often with smaller distances between different classes, suggesting potential classification difficulties. This visual evidence aligns with our quantitative results, underscoring our model's ability to effectively integrate both image and text modalities for superior feature learning and class separation in RSSC tasks.

\textbf{Heatmaps for Ablation Model Comparison: }Fig.~\ref{fig:heat} presents examples of attention heatmaps for our proposed model and the ablation models: \textit{No CAtt}, \textit{ICAtt}, and \textit{TCAtt}. These heatmaps illustrate the regions of focus for each model when processing and classifying remote sensing scenes.

From the heatmap comparison, it is evident that our proposed method surpasses the ablation models in identifying relevant attention areas. Specifically, when comparing the \textit{No CAtt} model with ours for the Viaduct class, we observe that the \textit{No CAtt} model's attention areas are broader and less focused, with lower contrast between high and low attention regions. This scattered attention pattern results from the absence of the cross-attention module, causing the model to rely solely on self-attention within the image modality, thus missing the guidance that textual information provides.

In contrast, the \textit{ICAtt} heatmap shows concentrated image-wise attention areas with high contrast, while text-related attention areas are more dispersed and less intense. This is because the \textit{ICAtt} model leverages textual information to guide the focus on image features, but the text features themselves remain less influenced by the image content. Meanwhile, the \textit{TCAtt} heatmaps display more concentrated and narrow text-related attention areas, which correspond to textual elements that enhance image understanding.

The heatmaps of our model, however, demonstrate that the dual-attention module allows for mutual reinforcement of attention in both text and image modalities. This results in more precise attention areas with high contrast between high and low-attention regions. Consequently, our model achieves accurate localization of important areas, with a strong correlation between the image and text modalities and the relevant class labels, which is quite essential in emphasizing areas-of-interest for RSSC tasks.

\textbf{Heatmaps for Proposed Model with Different Text Sources:} Fig.~\ref{fig:ucm} illustrates the differences in heatmaps generated using human-annotated captions versus VLM-generated descriptions. The model's attention areas are significantly influenced by the varying levels of detail and contextual information provided by each text source.

Human-annotated captions are typically more concise, focusing on specific elements within the image, such as ``intersection," ``house," and ``plants" in the intersection image (Fig.~\ref{fig:ucm}a), or ``villa," ``plants," and ``cars" in the sparse residential image (Fig.~\ref{fig:ucm}d). This targeted focus often results in more precise and localized attention areas. In contrast, VLM-generated descriptions tend to be more detailed, offering additional background information like ``multiple lanes of traffic both moving and stationary" or ``green vegetation, a building with a black roof, and a paved area." This increased detail encourages the model to adopt a broader attention span, potentially highlighting a wider range of areas within the image.

The diversity of descriptions provided by VLMs can enhance the model's ability to generate better representations, particularly in challenging scenarios. Moreover, these varied descriptions provide context that helps the model recognize relationships between different elements, thereby highlighting the most relevant areas for classification.

This is evident in the heatmaps, where areas of interest show high contrast compared to less important regions, indicating an effective focus for classification. The broader and more detailed context provided by VLM-generated descriptions aids in identifying and distinguishing key features within complex scenes.

\section{Further Discussion}

Beyond our experiments, we would like to explore potential applications of our model, such as zero-shot classification. We aim to experimentally demonstrate that the multimodal representation learned by our model, enriched by knowledge from VLMs, is advantageous in establishing a mechanism for knowledge transfer from known to unknown categories.

\begin{figure}[!htb]
\begin{center}

		\includegraphics[width=1.0\columnwidth]{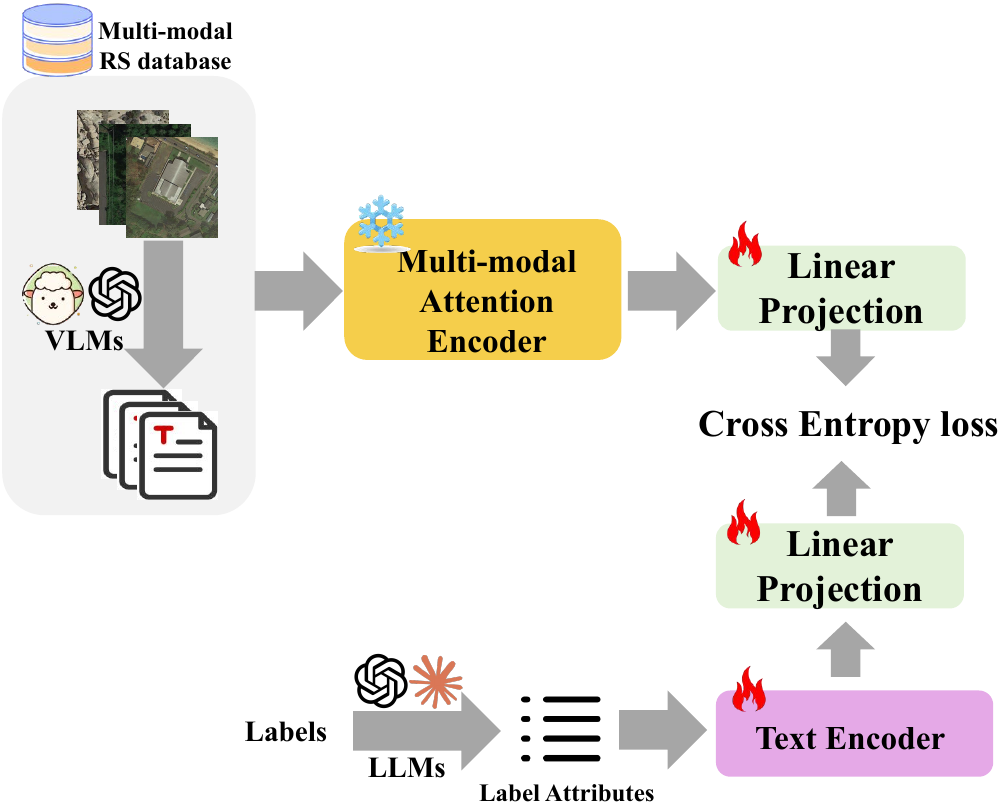}
	\caption{Illustration of the zero-shot learning framework. }
\label{fig:zeroshot}
\end{center}
\end{figure}

The constructed zero-shot model is illustrated in Fig.~\ref{fig:zeroshot}. We first build a multimodal remote sensing (RS) database, which includes images of all classes along with their corresponding VLM-generated descriptions. We also use large language models (LLMs) to expand the description labels, obtaining a comprehensive set of label attributes. For example, the expanded label attribute for the ``Airport" category might include: 

\texttt{A variety of elements such as runways and taxiways where planes take off and land, terminals bustling with passengers, and control towers overseeing operations. There can also be hangars housing aircraft, extensive parking lots, fuel storage facilities, radar equipment, and areas where baggage is handled.}

The multimodal RS database and the label attribute set are divided into known and unknown sets separately. The known set is used for training, while the unknown set is reserved for testing. In zero-shot models, no data from the unknown set is exposed during training. During the training phase, data from the multimodal RS database is processed through our model, which was pre-trained on Million-AID level-3 dataset and frozen (with the projection layer removed). The processed data is then passed through a trainable linear layer. Simultaneously, the corresponding label attributes are processed using a trainable CLIP text encoder to obtain their representations. The multimodal representations and the label attribute representations are paired and then used to calculate the similarity where the Cross Entropy loss is optimized. The objective during the zero-shot model training is to maximize the similarity between the multimodal RS data and the corresponding label attributes from the same class. Once the training is complete, the performance of the model is tested on the unknown classes to evaluate its zero-shot inference capabilities. 

\begin{table}[]
\centering
\renewcommand{\arraystretch}{1.5}
\caption{Comparison of experimental results on AID dataset for image- and multimodal-based zero-shot inference in terms of top-1 Accuracy (\%).}
\label{table:zero}
 \resizebox{0.36\textwidth}{!}{
\begin{tabular}{c|c|c|c}
\hline
Split Ratio              & 25/5   & 20/10   & 15/15 \\ \hline \hline
Image-based              & 75.4 & 54.5 & 49.6  \\ \hline
Multimodal-based         & 78.0 & 58.5 & 52.0  \\ \hline
\end{tabular}
}
\end{table}

We conducted experiments on the AID dataset, where the 30 categories were randomly divided into three different Known/Unknown ratios: 25/5, 20/10, and 15/15. In these ratios, for example, 25/5 means that 25 categories are treated as Known classes, and the remaining 5 are considered Unknown classes. For the zero-shot baseline, we replaced both the multimodal RS database and our model with pure images and image-only (\textit{IO}) model. The results obtained from these experiments are shown in Table.~\ref{table:zero}, where top-1 accuracy are reported. It is demonstrated that multimodal representation from our model enhances the zero-shot model's ability to bridge the variance between known and unknown classes compared to zero-shot model that relies solely on image data. This suggests that leveraging additional data modalities can be crucial for improving classification accuracy in zero-shot learning scenarios. By supplementing the visual information with textual modalities, our model's representation can better capture shared attributes or semantic relationships, effectively bridging the gap between known and unknown categories.

\section{Conclusion and Future Work}
In this study, we proposed a VLM-based multimodal dual-attention framework for remote sensing scene classification. Through a novel multimodal dual-attention module, our model effectively leverages the complementary strengths of visual images and textual data from VLM-generated text descriptions. Through comprehensive experimental evaluation and comparative analysis against both unimodal baselines and ablation multimodal fusion approaches, we demonstrate the superior performance of our architecture. \textcolor{black}{Our findings establish a robust methodology for integrating large-scale VLMs into remote sensing applications.} 

Our supplementary experiments examined the relative impact of different textual sources on model performance, revealing that VLM-generated descriptions outperform human-annotated captions. Furthermore, we demonstrated the framework's capability for zero-shot classification, highlighting the potential of robust multimodal representations in novel classification scenarios.


The success of our framework opens up new avenues for integrating large models and multimodal fusion learning in remote sensing scene classification and other applications. Looking ahead, we plan to extend our work to explore image-text retrieval applications, bringing our approach closer to real-world scenarios. \textcolor{black}{Additionally, our future research could investigate the integration of large models with other data modalities, such as temporal or spectral information, as well as extend the framework to more complex tasks such as segmentation and dense prediction.}

\newpage

\bibliography{main}

\begin{thebibliography}{10}
\providecommand{\url}[1]{#1}
\csname url@samestyle\endcsname
\providecommand{\newblock}{\relax}
\providecommand{\bibinfo}[2]{#2}
\providecommand{\BIBentrySTDinterwordspacing}{\spaceskip=0pt\relax}
\providecommand{\BIBentryALTinterwordstretchfactor}{4}
\providecommand{\BIBentryALTinterwordspacing}{\spaceskip=\fontdimen2\font plus
\BIBentryALTinterwordstretchfactor\fontdimen3\font minus \fontdimen4\font\relax}
\providecommand{\BIBforeignlanguage}[2]{{%
\expandafter\ifx\csname l@#1\endcsname\relax
\typeout{** WARNING: IEEEtran.bst: No hyphenation pattern has been}%
\typeout{** loaded for the language `#1'. Using the pattern for}%
\typeout{** the default language instead.}%
\else
\language=\csname l@#1\endcsname
\fi
#2}}
\providecommand{\BIBdecl}{\relax}
\BIBdecl

\bibitem{thapa2023deep}
A.~Thapa, T.~Horanont, B.~Neupane, and J.~Aryal, ``Deep learning for remote sensing image scene classification: A review and meta-analysis,'' \emph{Remote Sensing}, vol.~15, no.~19, p. 4804, 2023.

\bibitem{oghaz2019scene}
M.~M.~D. Oghaz, M.~Razaak, H.~Kerdegari, V.~Argyriou, and P.~Remagnino, ``Scene and environment monitoring using aerial imagery and deep learning,'' in \emph{2019 15th International Conference on Distributed Computing in Sensor Systems (DCOSS)}.\hskip 1em plus 0.5em minus 0.4em\relax IEEE, 2019, pp. 362--369.

\bibitem{fang2022spatial}
F.~Fang, L.~Zeng, S.~Li, D.~Zheng, J.~Zhang, Y.~Liu, and B.~Wan, ``Spatial context-aware method for urban land use classification using street view images,'' \emph{ISPRS Journal of Photogrammetry and Remote Sensing}, vol. 192, pp. 1--12, 2022.

\bibitem{su2021urban}
Y.~Su, Y.~Zhong, Q.~Zhu, and J.~Zhao, ``Urban scene understanding based on semantic and socioeconomic features: From high-resolution remote sensing imagery to multi-source geographic datasets,'' \emph{ISPRS Journal of Photogrammetry and Remote Sensing}, vol. 179, pp. 50--65, 2021.

\bibitem{gu2019survey}
Y.~Gu, Y.~Wang, and Y.~Li, ``A survey on deep learning-driven remote sensing image scene understanding: Scene classification, scene retrieval and scene-guided object detection,'' \emph{Applied sciences}, vol.~9, no.~10, p. 2110, 2019.

\bibitem{dos2010evaluating}
J.~A. dos Santos, O.~A. Penatti, and R.~d.~S. Torres, ``Evaluating the potential of texture and color descriptors for remote sensing image retrieval and classification,'' in \emph{International conference on computer vision theory and applications}, vol.~2.\hskip 1em plus 0.5em minus 0.4em\relax SCITEPRESS, 2010, pp. 203--208.

\bibitem{luo2013indexing}
B.~Luo, S.~Jiang, and L.~Zhang, ``Indexing of remote sensing images with different resolutions by multiple features,'' \emph{IEEE Journal of Selected Topics in Applied Earth Observations and Remote Sensing}, vol.~6, no.~4, pp. 1899--1912, 2013.

\bibitem{sheng2012high}
G.~Sheng, W.~Yang, T.~Xu, and H.~Sun, ``High-resolution satellite scene classification using a sparse coding based multiple feature combination,'' \emph{International journal of remote sensing}, vol.~33, no.~8, pp. 2395--2412, 2012.

\bibitem{cheng2020remote}
G.~Cheng, X.~Xie, J.~Han, L.~Guo, and G.-S. Xia, ``Remote sensing image scene classification meets deep learning: Challenges, methods, benchmarks, and opportunities,'' \emph{IEEE Journal of Selected Topics in Applied Earth Observations and Remote Sensing}, vol.~13, pp. 3735--3756, 2020.

\bibitem{liu2018scene}
Y.~Liu, Y.~Zhong, and Q.~Qin, ``Scene classification based on multiscale convolutional neural network,'' \emph{IEEE Transactions on Geoscience and Remote Sensing}, vol.~56, no.~12, pp. 7109--7121, 2018.

\bibitem{guo2020gan}
D.~Guo, Y.~Xia, and X.~Luo, ``Gan-based semisupervised scene classification of remote sensing image,'' \emph{IEEE Geoscience and Remote Sensing Letters}, vol.~18, no.~12, pp. 2067--2071, 2020.

\bibitem{lv2022scvit}
P.~Lv, W.~Wu, Y.~Zhong, F.~Du, and L.~Zhang, ``Scvit: A spatial-channel feature preserving vision transformer for remote sensing image scene classification,'' \emph{IEEE Transactions on Geoscience and Remote Sensing}, vol.~60, pp. 1--12, 2022.

\bibitem{bi2022vision}
M.~Bi, M.~Wang, Z.~Li, and D.~Hong, ``Vision transformer with contrastive learning for remote sensing image scene classification,'' \emph{IEEE Journal of Selected Topics in Applied Earth Observations and Remote Sensing}, vol.~16, pp. 738--749, 2022.

\bibitem{shendryk2018deep}
I.~Shendryk, Y.~Rist, R.~Lucas, P.~Thorburn, and C.~Ticehurst, ``Deep learning-a new approach for multi-label scene classification in planetscope and sentinel-2 imagery,'' in \emph{IGARSS 2018-2018 IEEE International Geoscience and Remote Sensing Symposium}.\hskip 1em plus 0.5em minus 0.4em\relax IEEE, 2018, pp. 1116--1119.

\bibitem{liu2023multi}
Q.~Liu, M.~He, Y.~Kuang, L.~Wu, J.~Yue, and L.~Fang, ``A multi-level label-aware semi-supervised framework for remote sensing scene classification,'' \emph{IEEE Transactions on Geoscience and Remote Sensing}, 2023.

\bibitem{he2018high}
Y.~He and Q.~Weng, \emph{High spatial resolution remote sensing: data, analysis, and applications}.\hskip 1em plus 0.5em minus 0.4em\relax CRC press, 2018.

\bibitem{yuan2021lightweight}
Z.~Yuan, W.~Zhang, X.~Rong, X.~Li, J.~Chen, H.~Wang, K.~Fu, and X.~Sun, ``A lightweight multi-scale crossmodal text-image retrieval method in remote sensing,'' \emph{IEEE Transactions on Geoscience and Remote Sensing}, vol.~60, pp. 1--19, 2021.

\bibitem{wang2022multi}
S.~Wang, X.~Ye, Y.~Gu, J.~Wang, Y.~Meng, J.~Tian, B.~Hou, and L.~Jiao, ``Multi-label semantic feature fusion for remote sensing image captioning,'' \emph{ISPRS Journal of Photogrammetry and Remote Sensing}, vol. 184, pp. 1--18, 2022.

\bibitem{cheng2023teaw}
K.~Cheng, C.~Yang, Z.~Fan, D.~Wu, and N.~Guan, ``Teaw: Text-aware few-shot remote sensing image scene classification,'' in \emph{ICASSP 2023-2023 IEEE International Conference on Acoustics, Speech and Signal Processing (ICASSP)}.\hskip 1em plus 0.5em minus 0.4em\relax IEEE, 2023, pp. 1--5.

\bibitem{liu2024remoteclip}
F.~Liu, D.~Chen, Z.~Guan, X.~Zhou, J.~Zhu, Q.~Ye, L.~Fu, and J.~Zhou, ``Remoteclip: A vision language foundation model for remote sensing,'' \emph{IEEE Transactions on Geoscience and Remote Sensing}, 2024.

\bibitem{tong2020land}
X.-Y. Tong, G.-S. Xia, Q.~Lu, H.~Shen, S.~Li, S.~You, and L.~Zhang, ``Land-cover classification with high-resolution remote sensing images using transferable deep models,'' \emph{Remote Sensing of Environment}, vol. 237, p. 111322, 2020.

\bibitem{shen2021dual}
J.~Shen, T.~Zhang, Y.~Wang, R.~Wang, Q.~Wang, and M.~Qi, ``A dual-model architecture with grouping-attention-fusion for remote sensing scene classification,'' \emph{Remote Sensing}, vol.~13, no.~3, p. 433, 2021.

\bibitem{wang2018scene}
Q.~Wang, S.~Liu, J.~Chanussot, and X.~Li, ``Scene classification with recurrent attention of vhr remote sensing images,'' \emph{IEEE Transactions on Geoscience and Remote Sensing}, vol.~57, no.~2, pp. 1155--1167, 2018.

\bibitem{alhichri2021classification}
H.~Alhichri, A.~S. Alswayed, Y.~Bazi, N.~Ammour, and N.~A. Alajlan, ``Classification of remote sensing images using efficientnet-b3 cnn model with attention,'' \emph{IEEE access}, vol.~9, pp. 14\,078--14\,094, 2021.

\bibitem{alhichri2023rs}
H.~Alhichri, ``Rs-deepsuperlearner: fusion of cnn ensemble for remote sensing scene classification,'' \emph{Annals of GIS}, vol.~29, no.~1, pp. 121--142, 2023.

\bibitem{hou2023contextual}
Y.-E. Hou, K.~Yang, L.~Dang, and Y.~Liu, ``Contextual spatial-channel attention network for remote sensing scene classification,'' \emph{IEEE Geoscience and Remote Sensing Letters}, 2023.

\bibitem{peng2023local}
T.~Peng, J.~Yi, and Y.~Fang, ``A local--global interactive vision transformer for aerial scene classification,'' \emph{IEEE Geoscience and Remote Sensing Letters}, vol.~20, pp. 1--5, 2023.

\bibitem{soroush2023nir}
R.~Soroush and Y.~Baleghi, ``Nir/rgb image fusion for scene classification using deep neural networks,'' \emph{The Visual Computer}, vol.~39, no.~7, pp. 2725--2739, 2023.

\bibitem{liu2021mrssc}
K.~Liu, A.~Wu, X.~Wan, and S.~Li, ``Mrssc: a benchmark dataset for multimodal remote sensing scene classification,'' \emph{The International Archives of the Photogrammetry, Remote Sensing and Spatial Information Sciences}, vol.~43, pp. 785--792, 2021.

\bibitem{ma2023optical}
Y.~Ma, J.~Pei, X.~Zhang, W.~Huo, Y.~Zhang, Y.~Huang, and J.~Yang, ``An optical image-aided approach for zero-shot sar image scene classification,'' in \emph{2023 IEEE Radar Conference (RadarConf23)}.\hskip 1em plus 0.5em minus 0.4em\relax IEEE, 2023, pp. 1--6.

\bibitem{sun2024alignment}
X.~Sun, J.~Gao, and Y.~Yuan, ``Alignment and fusion using distinct sensor data for multimodal aerial scene classification,'' \emph{IEEE Transactions on Geoscience and Remote Sensing}, 2024.

\bibitem{zhang2023text2seg}
J.~Zhang, Z.~Zhou, G.~Mai, M.~Hu, Z.~Guan, S.~Li, and L.~Mu, ``Text2seg: Remote sensing image semantic segmentation via text-guided visual foundation models,'' \emph{arXiv preprint arXiv:2304.10597}, 2023.

\bibitem{yuan2023parameter}
Y.~Yuan, Y.~Zhan, and Z.~Xiong, ``Parameter-efficient transfer learning for remote sensing image-text retrieval,'' \emph{IEEE Transactions on Geoscience and Remote Sensing}, 2023.

\bibitem{tang2023interacting}
X.~Tang, Y.~Wang, J.~Ma, X.~Zhang, F.~Liu, and L.~Jiao, ``Interacting-enhancing feature transformer for cross-modal remote-sensing image and text retrieval,'' \emph{IEEE Transactions on Geoscience and Remote Sensing}, vol.~61, pp. 1--15, 2023.

\bibitem{qiu2024few}
C.~Qiu, X.~Zhang, X.~Tong, N.~Guan, X.~Yi, K.~Yang, J.~Zhu, and A.~Yu, ``Few-shot remote sensing image scene classification: Recent advances, new baselines, and future trends,'' \emph{ISPRS Journal of Photogrammetry and Remote Sensing}, vol. 209, pp. 368--382, 2024.

\bibitem{byra2023few}
M.~Byra, M.~F. Rachmadi, and H.~Skibbe, ``Few-shot medical image classification with simple shape and texture text descriptors using vision-language models,'' \emph{arXiv preprint arXiv:2308.04005}, 2023.

\bibitem{ji2024vision}
J.~Ji, Y.~Hou, X.~Chen, Y.~Pan, and Y.~Xiang, ``Vision-language model for generating textual descriptions from clinical images: model development and validation study,'' \emph{JMIR Formative Research}, vol.~8, p. e32690, 2024.

\bibitem{qin2022medical}
\BIBentryALTinterwordspacing
Z.~Qin, H.~Yi, Q.~Lao, and K.~Li, ``Medical image understanding with pretrained vision language models: A comprehensive study,'' 2023. [Online]. Available: \url{https://openreview.net/forum?id=txlWziuCE5W}
\BIBentrySTDinterwordspacing

\bibitem{zhong2024vlm}
L.~Zhong, X.~Liao, S.~Zhang, X.~Zhang, and G.~Wang, ``Vlm-cpl: Consensus pseudo labels from vision-language models for human annotation-free pathological image classification,'' \emph{arXiv preprint arXiv:2403.15836}, 2024.

\bibitem{saha2024improved}
O.~Saha, G.~Van~Horn, and S.~Maji, ``Improved zero-shot classification by adapting vlms with text descriptions,'' in \emph{Proceedings of the IEEE/CVF Conference on Computer Vision and Pattern Recognition}, 2024, pp. 17\,542--17\,552.

\bibitem{tzelepi2024exploiting}
M.~Tzelepi and V.~Mezaris, ``Exploiting lmm-based knowledge for image classification tasks,'' in \emph{International Conference on Engineering Applications of Neural Networks}.\hskip 1em plus 0.5em minus 0.4em\relax Springer, 2024, pp. 166--177.

\bibitem{liu2024visual}
H.~Liu, C.~Li, Q.~Wu, and Y.~J. Lee, ``Visual instruction tuning,'' \emph{Advances in neural information processing systems}, vol.~36, 2024.

\bibitem{dosovitskiy2020image}
A.~Dosovitskiy, L.~Beyer, A.~Kolesnikov, D.~Weissenborn, X.~Zhai, T.~Unterthiner, M.~Dehghani, M.~Minderer, G.~Heigold, S.~Gelly \emph{et~al.}, ``An image is worth 16x16 words: Transformers for image recognition at scale,'' \emph{arXiv preprint arXiv:2010.11929}, 2020.

\bibitem{radford2021learning}
A.~Radford, J.~W. Kim, C.~Hallacy, A.~Ramesh, G.~Goh, S.~Agarwal, G.~Sastry, A.~Askell, P.~Mishkin, J.~Clark \emph{et~al.}, ``Learning transferable visual models from natural language supervision,'' in \emph{International conference on machine learning}.\hskip 1em plus 0.5em minus 0.4em\relax PMLR, 2021, pp. 8748--8763.

\bibitem{kirillov2023segment}
A.~Kirillov, E.~Mintun, N.~Ravi, H.~Mao, C.~Rolland, L.~Gustafson, T.~Xiao, S.~Whitehead, A.~C. Berg, W.-Y. Lo \emph{et~al.}, ``Segment anything,'' in \emph{Proceedings of the IEEE/CVF International Conference on Computer Vision}, 2023, pp. 4015--4026.

\end{thebibliography}
\bibliographystyle{IEEEtran}

\vfill

\end{document}